\pdfoutput=1

\documentclass[11pt]{article}

\usepackage[preprint]{acl}

\usepackage{times}
\usepackage{latexsym}

\usepackage[T1]{fontenc}

\usepackage[utf8]{inputenc}

\usepackage{microtype}

\usepackage{inconsolata}

\usepackage{graphicx}

\usepackage{CJKutf8}
\usepackage{float}
\usepackage{adjustbox}
\usepackage{booktabs}
\usepackage{hyperref}
\usepackage{lipsum}
\usepackage{tabularx}
\usepackage{makecell}

\newcommand{\modified}[1]{{#1}}
\usepackage{subcaption}

\title{Improbable Bigrams Expose Vulnerabilities of Incomplete Tokens in Byte-Level Tokenizers}

\author{
    \textbf{Eugene Jang\textsuperscript{1\thanks{\hspace{0.05cm}Work performed while at S2W Inc.}}} \hspace{1em}
    \textbf{Kimin Lee\textsuperscript{2}} \hspace{1em}
    \textbf{Jin-Woo Chung\textsuperscript{3}} \hspace{1em}
    \textbf{Keuntae Park\textsuperscript{3}} \hspace{1em} 
    \textbf{Seungwon Shin\textsuperscript{2}}
    \vspace{0.1cm} \\
    \textsuperscript{1}Northeastern University\hspace{1em}
    \textsuperscript{2}KAIST\hspace{1em}
    \textsuperscript{3}S2W Inc.\vspace{0.1cm}\\
    \textsuperscript{1}\texttt{jang.eu@northeastern.edu}\hspace{1em}
    \textsuperscript{2}\texttt{\{kiminlee,claude\}@kaist.ac.kr}\\
    \textsuperscript{3}\texttt{\{jwchung,keuntae.park\}@s2w.inc}
}

\begin{document}
\maketitle
\begin{abstract}
Tokenization is a crucial step that bridges human-readable text with model-readable discrete tokens. 
However, recent studies have revealed that tokenizers can be exploited to elicit unwanted model behaviors. 
In this work, we investigate \textit{incomplete tokens}, i.e., undecodable tokens with stray bytes resulting from byte-level byte-pair encoding (BPE) tokenization. 
We hypothesize that such tokens are heavily reliant on their adjacent tokens and are fragile when paired with unfamiliar tokens. 
To demonstrate this vulnerability, we introduce \textit{improbable bigrams}: out-of-distribution combinations of incomplete tokens designed to exploit their dependency. 
Our experiments show that improbable bigrams are significantly prone to hallucinatory behaviors.
\modified{Surprisingly, the same phrases have drastically lower rates of hallucination (90\% reduction in Llama3.1) when an alternative tokenization is used.}
We caution against the potential vulnerabilities introduced by byte-level BPE tokenizers, which may \modified{introduce blind spots to language models.}

\end{abstract}
\section{Introduction}

Tokenization is an important step in the large language model (LLM) pipeline, serving as the bridge between text inputs and the discrete tokens processed by the model.
Improper tokenization can lead to undesirable behaviors, such as glitch token hallucinations~\cite{rumbelow2023solidgoldmagikarp, land2024fishingmagikarpautomaticallydetecting} and errors in numerical reasoning~\cite{singh2024tokenizationcountsimpacttokenization}.
Furthermore, it has been observed that tokenizers can introduce elements of unfairness~\cite{petrov2023token_unfairness} and bias~\cite{ovalle-etal-2024-tokenization} into models.
Given the increasing duration and cost of model training, investigating and understanding the impact of the tokenizer is becoming increasingly important.


\begin{figure}[t!]
    \centering
    \includegraphics[width=1.04\columnwidth]{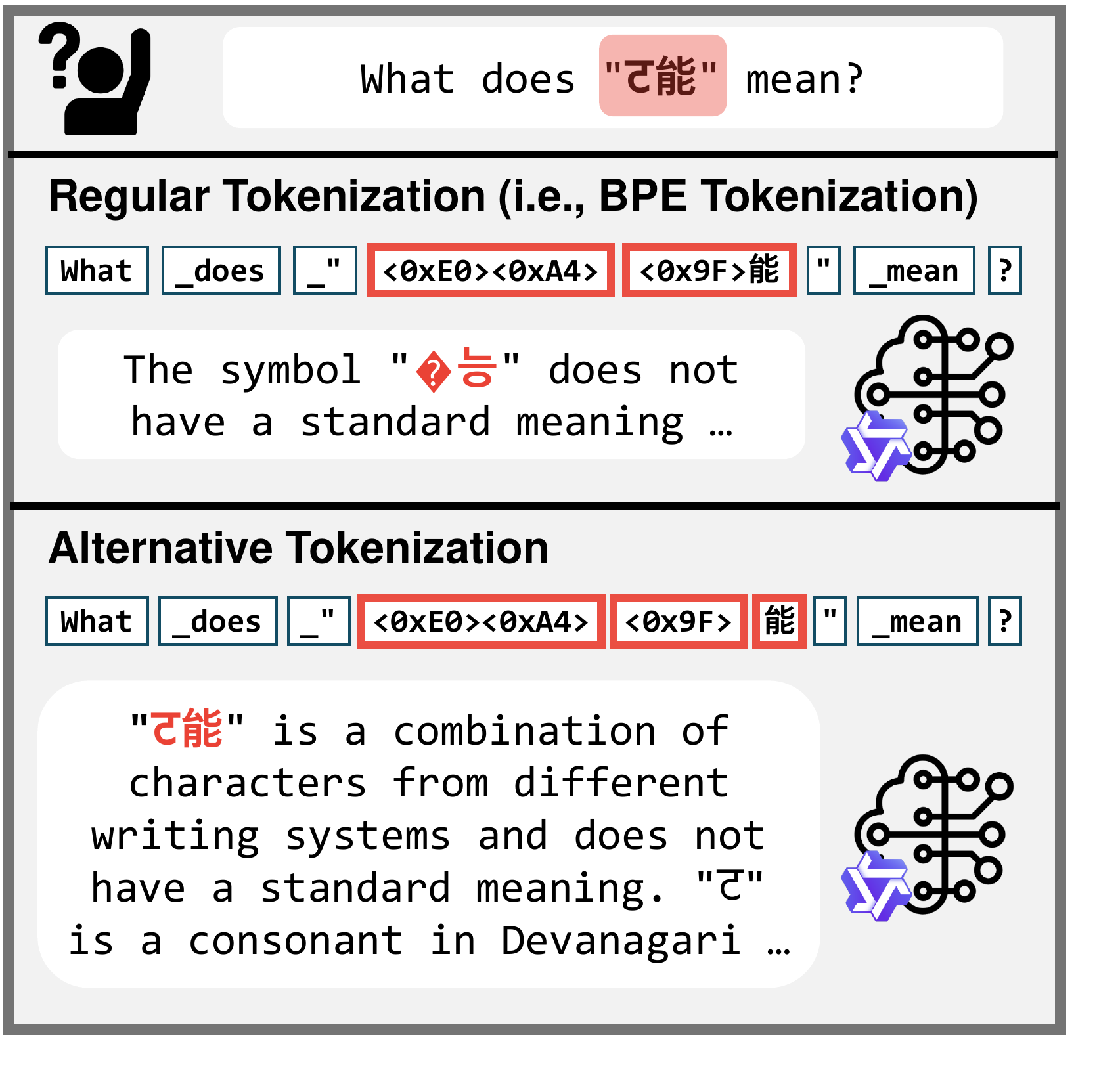}
    \vspace{-0.35in}
    \caption{An improbable bigram phrase that combines two incomplete tokens to cause hallucinatory behaviors in the Qwen2.5 model. This behavior persists across multiple models and with well-trained tokens. An alternative tokenization of the same phrase does not cause hallucinations.}
    \label{fig:Fig1}
    \vspace{-0.2in}
\end{figure}

Recent works specifically analyze the tokenization step to identify inputs that provoke unwanted model behaviors. 
\citet{land2024fishingmagikarpautomaticallydetecting} propose embedding layer heuristics to identify {\em glitch} tokens --- undertrained tokens that cause hallucinations and also enable jailbreaks~\cite{geiping2024coercingllmsrevealalmost}.
\citet{wang2024tokenizationmattersdegradinglarge} create adversarial questions designed to induce incorrect segmentation by the tokenizer, which degrades model performance.

In this work, we investigate a specific vulnerability associated with \textit{incomplete tokens} in byte-level byte pair encoding (BPE) tokenizers. 
These tokens, also known as undecodable tokens, are byte-level tokens that cannot be decoded independently and must appear in conjunction with certain other tokens to form legal Unicode characters.
We explore the fragility of these tokens by constructing \textit{improbable bigrams}, which are unlikely yet permissible combinations of two incomplete tokens.
Similar to glitch tokens, improbable bigrams cause hallucinations to benign user requests, as seen in Figure~\ref{fig:Fig1}.
However, unlike glitch tokens, improbable bigrams can cause hallucinatory behavior even when their constituent tokens are well-trained.

Our experiments across multiple LLM families (such as Llama-3.1, Qwen2.5, and Mistral-Nemo) demonstrate that these bigrams are significantly more prone to producing hallucinations compared to bigrams formed from complete tokens.
Intriguingly, alternative tokenizations of the same phrases rarely exhibit hallucinations.
We believe that our findings contribute to the growing body of research on understanding the potential weaknesses of BPE tokenization, with implications for the development of more robust language models.

\section{BPE Tokenization}
\begin{table}
  \centering
  \resizebox{\columnwidth}{!}{%
  \begin{tabular}{lccc}
    \hline
    \thead{\textbf{Tokenizer}} & \thead{\textbf{Vocab}\\\textbf{Size}} & \thead{\textbf{Incomplete}\\\textbf{Tokens}} & \thead{\textbf{Incomplete}\\\textbf{Bigrams}}\\
    
    \hline
    Meta-Llama-3.1     & {128k} &   {1224}&{71k}      \\
    Exaone-3.0     & {102k}     &   {1222}&{36k}    \\
    Qwen2.5     & {151k}       &   {1320}&{39k}     \\
    Mistral-Nemo     & {131k}   &   {1307}&{135k}         \\
    Command-R-v01      & {255k}&   {2956}&{1479k}            \\
    \hline
  \end{tabular}%
  }
  \caption{Vocabulary sizes and number of incomplete tokens in different tokenizers. Incomplete bigrams refer to the total number of legal bigrams that can be created by combining two incomplete tokens.}
  \label{tab:stats}
\end{table}

Byte pair encoding (BPE), originally developed as a data compression algorithm~\cite{Gage1994ANA}, has evolved into a popular tokenization scheme used by many modern language models~\cite{sennrich-etal-2016-neural}. BPE tokenization iteratively expands the vocabulary by selecting frequent pairs of existing tokens as new vocabulary items. 
In particular, byte-level BPE, which applies the BPE algorithm at the byte-level rather than the character level~\cite{Wang2019NeuralMT}, offers distinct advantages. Firstly, byte-level tokenization eliminates out-of-vocabulary issues by representing all Unicode characters (154,998 in Unicode v16.0) as combinations of 256 base bytes, ensuring comprehensive coverage. Secondly, it allows for more efficient data compression. The widespread adoption of byte-level tokenization in cutting-edge models such as GPT-4 and Llama 3.1 underscores its effectiveness.


While some have attributed BPE's effectiveness to its compression~\cite{galle-2019-investigating, goldman-etal-2024-unpacking}, recent work has challenged this assumption~\cite{schmidt2024tokenizationcompression}.
\citet{bostrom-durrett-2020-byte} point out that greedy compression prioritizes frequency over linguistically meaningful boundaries.
This observation is supported by other works that report model improvements from morphologically motivated modifications to BPE tokenizers~\cite{limisiewicz-etal-2024-myte, bauwens-delobelle-2024-bpe}.
These works demonstrate that incorrect segmentation by BPE tokenization cause suboptimal model performance.


\section{Incomplete Tokens and Improbable Bigrams}
\label{sec:bigramsexplain}
\begin{CJK*}{UTF8}{gbsn}

Byte-level BPE tokenizers are agnostic not only to morphemes but also to character boundaries. 
Training with the BPE algorithm on texts where characters are encoded with multiple bytes \modified{(e.g., Cyrillic (2 bytes), CJK Unified Ideographs (3 bytes))} often results in the formation of \textit{incomplete tokens}.
Incomplete tokens contain stray bytes (like <0x9F> in the token `<0x9F>能') that do not correspond to recognizable characters.
These incomplete tokens constitute a small but significant portion of the vocabulary \modified{in byte-level models.}
Table~\ref{tab:stats} presents the number of discovered incomplete tokens in the vocabularies of various models employing byte-level tokenizers (see Appendix~\ref{sec:appendix-construction} for details).
\end{CJK*}

\newcolumntype{P}[1]{>{\ttfamily\raggedright\arraybackslash}p{#1}}
\begin{table*}[ht]
\small
\centering
\begin{tabular}{l | P{0.85\textwidth}}
\toprule 
\textbf{Task} & \multicolumn{1}{l}{\textbf{User prompt}} \\

\midrule
Repeat  &  Repeat this phrase exactly: ``\textbf{\{Phrase\}}''  \\
\midrule
Definition  &  What does ``\textbf{\{Phrase\}}'' mean?  \\
\midrule
Do You Know? &  Today I heard about ``\textbf{\{Phrase\}}''. Do you know what this means?    \\
\midrule  
Usernames &  \adjustbox{max width=0.86\textwidth, valign=m}{%
    \includegraphics{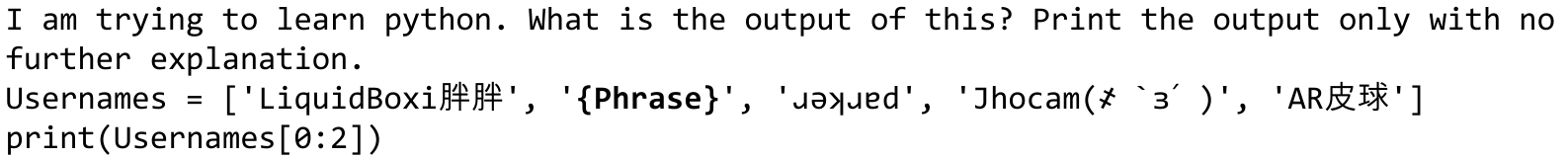}
}\\
\bottomrule
\end{tabular}
\caption{Prompt templates used to test phrase-level hallucinations of target phrases. At test time, \textbf{\{Phrase\}} is replaced with the tested phrase. If the model fails to repeat the phrase in \textit{all} four prompts, we consider the phrase to be hallucinatory.}
\label{tab:prompts}
\end{table*}
\begin{figure}[t!]
    \centering
    \includegraphics[width=1.1\columnwidth]{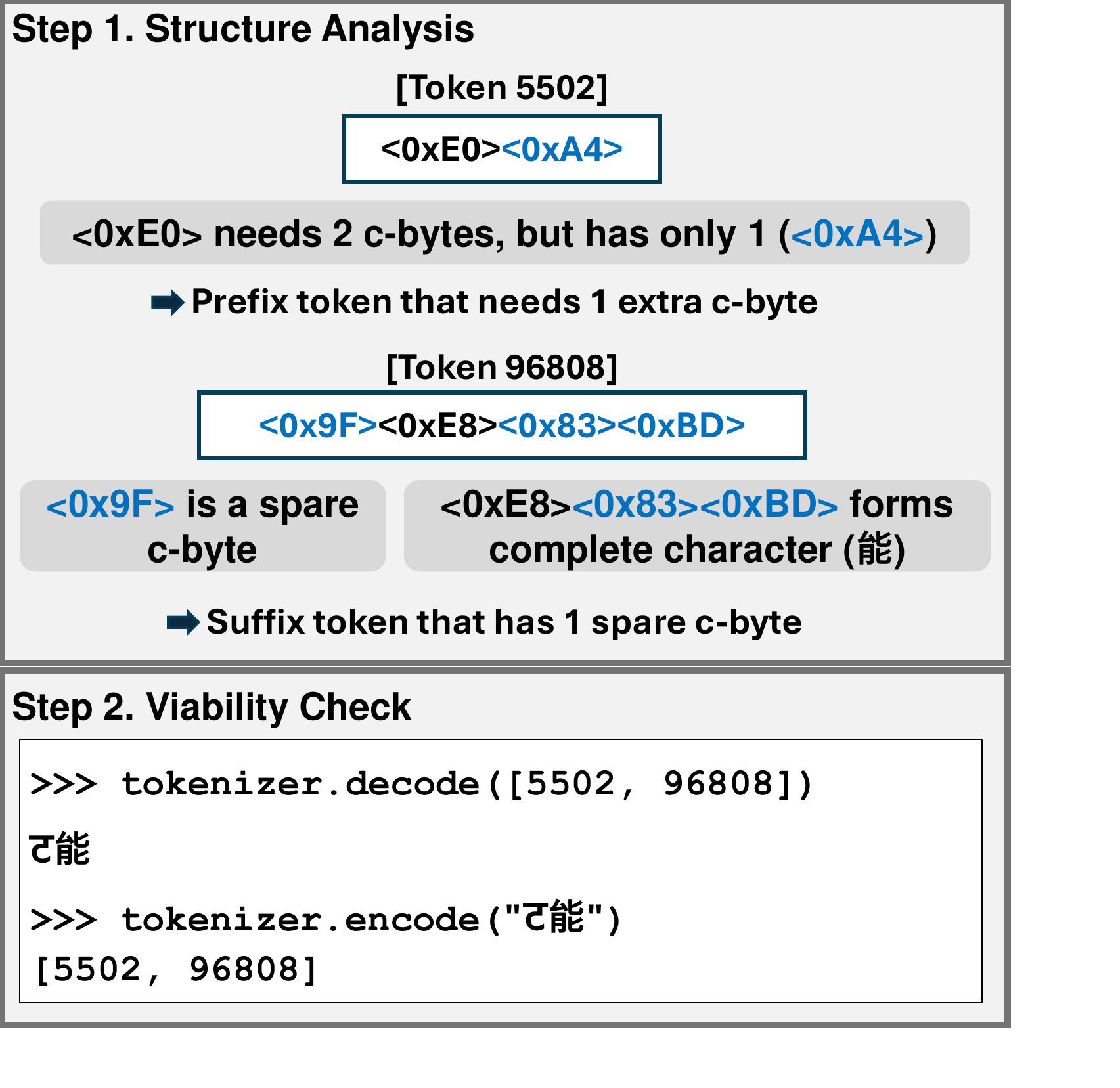}
    \caption{Incomplete tokens are analyzed for their structure to find potential bigram candidates. C-bytes (continuation bytes) are colored in \textcolor{darkblue}{blue}. Tokens with compatible structures are tested for viability and reachability through an decode-encode check.}
    \label{fig:Bigram}
\end{figure}

While incomplete tokens are indistinguishable from complete tokens from the model's perspective during training, their structure can influence how they are trained in subtle ways.
First, incomplete tokens can only occur alongside other incomplete or single-byte tokens, as the ``stray bytes'' of incomplete tokens require additional bytes to form a viable character.
This restriction limits the range of tokens they can be associated with.
Second, the potential characters that stray bytes can form into may have little to no semantic overlap, leading to a representation that is inherently ambiguous and heavily reliant on neighboring tokens for disambiguation. 
We hypothesize that these factors exacerbate one another, causing incomplete tokens to become overly dependent on their context and brittle in response to unfamiliar adjacent tokens.



\modified{To demonstrate this vulnerability, we create phrases that exploit incomplete tokens to cause hallucinations.}
First, we analyze the structure of incomplete tokens, categorizing those that end with stray bytes as {\em prefixes} and those that begin with stray bytes as {\em suffixes}. 
We further label prefixes and suffixes by the number of stray bytes they require to form a complete character, to find potential bigram combinations.
We create adversarial combinations of these tokens, termed \textit{improbable bigrams}, as illustrated in Figure~\ref{fig:Bigram}. 
These bigrams exploit the dependency of two incomplete tokens to complete each other. 
We note that not all combinations are viable due to tokenization rules and merge priorities. 
Table~\ref{tab:stats} lists the number of all possible legal bigrams created from incomplete tokens across different tokenizers.
\modified{To identify bigrams that are particularly out-of-distribution, we use the heuristic of multilinguality.
When a bigram forms a string, we check the Unicode script type of each character.
If the string contains characters from different Unicode scripts, we assume it is highly unlikely to have been encountered during training.}
In our experiments, we demonstrate that \textit{improbable bigrams} are prone to hallucinations.


\section{Hallucination Experiments}
\label{sect: hall-exp}
\vspace{0.1cm}
\noindent\textbf{Models} We experiment using five recently released instruction-trained models that use byte-level BPE tokenization: Meta-Llama-3.1-8B-Instruct\footnote{https://huggingface.co/meta-llama/Meta-Llama-3.1-8B-Instruct}, EXAONE-3.0.-7.8B-Instruct\footnote{https://huggingface.co/LGAI-EXAONE/EXAONE-3.0-7.8B-Instruct}, Qwen2.5-32B-Instruct\footnote{https://huggingface.co/Qwen/Qwen2.5-32B-Instruct}, Mistral-Nemo-Instruct-2407\footnote{https://huggingface.co/mistralai/Mistral-Nemo-Instruct-2407}, and C4AI-Command-R-v01\footnote{https://huggingface.co/CohereForAI/c4ai-command-r-v01}.
We use greedy decoding for all experiments.

\begin{table}[t]
\centering
\small
\begin{adjustbox}{width=\columnwidth}
\begin{tabular}{l|cc}
\toprule
\textbf{Models} & \thead{\textbf{Improbable}\\\textbf{Bigrams}} & \thead{\textbf{Baseline}\\\textbf{Bigrams}}\\
\midrule
Llama 3.1 & 48/100 (48\%) & 0/100 (0\%) \\
Exaone & 77/100 (77\%) & 20/100 (20\%) \\
Qwen2.5 & 33/100 (33\%) & 0/100 (0\%) \\
Mistral-Nemo & 52/71 (73\%) & 1/71 (1\%) \\
Command-R & 49/100 (49\%) & 8/100 (8\%) \\
\bottomrule
\end{tabular}
\end{adjustbox}
\caption{Number of hallucinations by \textbf{improbable bigram} phrases (with incomplete tokens) and by \textbf{baseline bigram} phrases (with complete tokens).}
\label{tab:hall_exps}
\end{table}

\noindent\textbf{Evaluation Prompts}
We assess each model's ability to accurately process prompts containing improbable bigram phrases. 
We use \modified{four} prompt templates designed to induce the model to repeat a target phrase from the input, regardless of the phrase's nonsensical or unusual nature, as detailed in Table~\ref{tab:prompts}.
\modified{
Our goal is to find phrases that models cannot process for any prompt.
Therefore, }we consider a phrase to have induced hallucinations only if the model fails to correctly repeat it across all \modified{four} prompt templates.


\vspace{0.1cm}
\noindent\textbf{Incomplete Token Selection} 
For each model, we construct improbable bigrams using the procedure outlined in Section~\ref{sec:bigramsexplain}.
Noting that undertrained tokens often lead to hallucinations, we focus our experiments on well-trained tokens to isolate confounding factors. 
To do this, we employ the embedding heuristics in \cite{land2024fishingmagikarpautomaticallydetecting} to detect whether a token in the vocabulary is undertrained.
Specifically, the heuristics estimate the degree of training for each token by taking L2 norms and cosine distances in the embedding matrix.
We use these metrics to order the entire vocabulary, and discard the lower (undertrained) half from our experiments.
From the upper half (well-trained half) of the vocabulary, we identify the incomplete tokens and create up to 100 improbable bigrams for each model.
\modified{Analysis of the languages used in the improbable bigrams (Appendix~\ref{sec:appendix-languages}) show language-pair distributions differ widely between models.}


\vspace{0.1cm}
\noindent\textbf{Baselines}
To better demonstrate the fragility of improbable bigrams, we create a baseline of bigrams constructed with complete tokens.
For each improbable bigram, we create a complete token counterpart by replacing the prefix and suffix with similarly-trained tokens.
That is, if the initial incomplete token is the \textit{i}-th token in the ranked vocabulary, we find a complete token that is slightly less trained (e.g. the (\textit{i-1})-th token).
This method ensures that each improbable bigram is compared to a baseline of complete bigrams with a similar level of undertraining.

\vspace{0.1cm}
\noindent\textbf{Results}
As shown in Table~\ref{tab:hall_exps}, improbable bigrams suffer from significantly higher rates of hallucination than their complete token counterparts across all tested models. 
This is distinct from hallucinations involving glitch tokens, as previously investigated by~\citet{land2024fishingmagikarpautomaticallydetecting}, which was attributed to undertrained tokens.
These results suggest that even well-trained incomplete tokens can struggle to faithfully represent textual inputs.

\section{\modified{Alternative Tokenization}}
\begin{table}[t]
\centering
\small
\begin{adjustbox}{width=\columnwidth}
\begin{tabular}{l|cc}
\toprule
\textbf{Models} &  \thead{\textbf{Incomplete}\\\textbf{Token}\\\textbf{Hallucinations}} & \thead{\textbf{Alternative}\\\textbf{Tokenization}\\\textbf{Hallucinations}}\\
\midrule
Llama 3.1 & 0.48 & 0.05 (↓90\%)\\
Exaone  & 0.77 & 0.50 (↓35\%)\\
Qwen2.5 &  0.33 & 0.12 (↓64\%)\\
Mistral-Nemo  & 0.73  & 0.01 (↓98\%) \\
Command-R & 0.49 & 0.55 \\
\bottomrule
\end{tabular}
\end{adjustbox}
\caption{Frequency of hallucinations from improbable bigram phrases with original tokenization (\textbf{Incomplete Tokens}) and presegmented tokenization (\textbf{Alternative Tokenization}). Reduction of hallucinations by alternative tokenization is denoted by ↓ when applicable.}
\label{tab:dehall_exps}
\end{table}

\modified{
Our previous experiments demonstrate that improbable bigram phrases are difficult for models to process.
We conduct further experiments to attribute this difficulty to the incomplete tokens.
We repeat the experiments, but pre-segment the target phrase to avoid character boundary-crossing tokenization.}
\begin{CJK*}{UTF8}{ipxm}
For instance, the improbable bigram phrase ``サーミ能'' is normally tokenized into incomplete tokens ``サー<0xE3><0x83>'' and ``<0x9F>能''. However, it can be presegmented to isolate the character formed from stray bytes (``ミ'').
We split the string into three parts (``サー'', ``ミ'', and ``能'') and tokenize each separately.
Afterwards, the tokens can be appended together to create an alternative token representation of the same phrase.
\end{CJK*}

\modified{
Compared to improbable bigrams, the alternative tokenization sequences can be suboptimal in two important ways.
First, these sequences cannot be generated by the tokenizer, ensuring that they are out-of-distribution for the model.
Second, the sequences consist of more than two tokens, requiring the model to recall more tokens correctly to repeat the phrase accurately.
}

\modified{
Table~\ref{tab:dehall_exps} shows results of the experiment repeated using the alternative tokenization scheme.
Despite the suboptimalities of alternative tokenization, most models, with the exception of Command-R (discussed in Appendix~\ref{sec:appendix-languages}) show significantly higher performance when alternative tokenization is used.
The same phrases tend to hallucinate significantly less when alternative tokenization is used, suggesting that the hallucinatory behaviors are being caused by the incomplete tokens.
}

\section{Discussion}
\vspace{0.1cm}
\noindent\textbf{Risks of Unrepeatability}
We have used a set of repetition-inducing prompts as a simple task that identifies phrases that cause model failures.
The repetition failures, stemming from incomplete tokens, are possibly indicative of other robustness failures linked to incomplete tokens.
However, in certain scenarios, non-repeatability itself can cause real world harms.
We highlight few such scenarios.
\begin{itemize}

\item Reliable Code and Data Handling: When models interact with code or databases, they must correctly preserve variable names or fixed values. If models are unable to preserve certain phrases, they can compromise the integrity of data.
\item Adversarial unrepeatability: Adversaries may exploit unrepeatable phrases to avoid intervention from LLM agents. For instance, an attacker can set their username to an unrepeatable phrases to evade a moderator agent.
\item Model Fingerprinting: Since improbable bigrams are model-specific due to vocabulary design, they could be used to fingerprint the architecture behind a closed or anonymized LLM service. This can facilitate sophisticated targeted attacks on private LLMs (e.g. adversarial triggers optimized for the target architecture). This can also create implications for model evaluations that rely on model anonymity (e.g. Chatbot Arena).
\end{itemize}

\vspace{0.1cm}
\noindent\textbf{Circumvention Strategies}
Our findings suggest that model developers should consider the consequences of vulnerable tokenizer design.
The vulnerability of incomplete tokens can be circumvented by preventing incomplete tokens from entering the final tokenizer vocabulary (before model training begins).
One method could involve vocabulary pruning~\cite{bauwens-delobelle-2024-bpe} after tokenizer training to remove incomplete tokens.
Another method would be to constrain BPE merges to respect character boundaries during tokenizer training~\cite{land2025bpestaysscriptstructured}.
If full Unicode coverage is not deemed necessary for a certain model, developers may opt to choose tokenization at the character-level before tokenizer-training.
We hope that demonstrating potential vulnerabilities of the default byte-level BPE can motivate more careful design of tokenizers.
\section{Conclusion}
Through improbable bigrams, we demonstrate vulnerabilities of incomplete tokens present in byte-level BPE tokenizers.
We conclude that incomplete tokens are significantly more prone to hallucinatory behaviors compared to complete tokens.
Our findings suggest that model developers must be mindful of potential unwanted behaviors caused by incomplete tokens.

\section*{Limitations}
This work provides evidence to suggest the fragility of incomplete tokens.
We assess phrase-level hallucinations as a proxy for larger potential harms of model misbehavior.
We limit our investigations to phrase-level hallucinations, rather than factual hallucinations.
While we occasionally observe that the models confidently assert incorrect explanations about what is likely non-existent terminology, this was not possible to evaluate systematically.
One reason is that the tested phrases contain scripts spanning many languages beyond the expertise of the authors.
Another is that the baseline levels of factual hallucination when asked about \textit{ordinary} terms in low-resource languages are already very high.
In pilot experiments, we found that even models with high conversational fluency often make mistakes in explaining non-existent terms.

Our findings on improbable bigrams causing hallucinations were consistent in all 5 models.
However, our secondary experiment on alternate tokenization did not cause improvements for the Cohere Command-R model.
It is difficult to isolate which design choices of the model contributes to this effect.
One hypothesis is that the model's significantly larger vocabulary size, which is approximately double the other studied models, may have impacted token training.
We have also conducted an analysis on the languages of the incomplete bigrams as a possible explanation, but found it inconclusive.
A more detailed investigation is left to future work.
There is a possibility that these findings may be influenced by other unknown and undocumented dependencies relating to tokenizer parameters, training parameters, and training data.

\bibliography{custom}

\appendix
\section{Incomplete Bigram Construction}
\label{sec:appendix-construction}

UTF-8 characters have a flexible number of bytes.
Each multi-byte character comprises of a starting byte and continuation bytes.
Starting bytes indicate how many bytes the character is composed of.
For example, the starting byte of a 3-byte character implies the following two characters will be continuation bytes.
A more detailed explanation of the UTF-8 protocol and byte structures can be found in~\cite{limisiewicz-etal-2024-myte}.

Using starting bytes and continuation bytes, we can identify whether a byte sequence requires more continuation bytes to be fulfilled or has excess continuation bytes.
As illustrated in Figure~\ref{fig:Bigram}, each incomplete token can be analyzed for their structure.
We test all possible combinations of incomplete bigrams by pairing bigrams of complementary structures.
We focus only on bigrams that users could use to induce the intended incomplete bigram structure.
Note that the resulting character of conjoined bytes may not always correspond to a usable Unicode character.
There is also a possibility that the resulting phrase is tokenized into a different sequence of tokens.
We test this by performing a decode-encode test to ensure the resulting string is valid and indeed tokenizes to the intended prefix and suffix.

\section{Role of Languages in Improbable Bigrams}
\label{sec:appendix-languages}

\begin{figure*}[t]
    \centering
    \begin{subfigure}[b]{\textwidth}
        \centering
        \includegraphics[width=\textwidth]{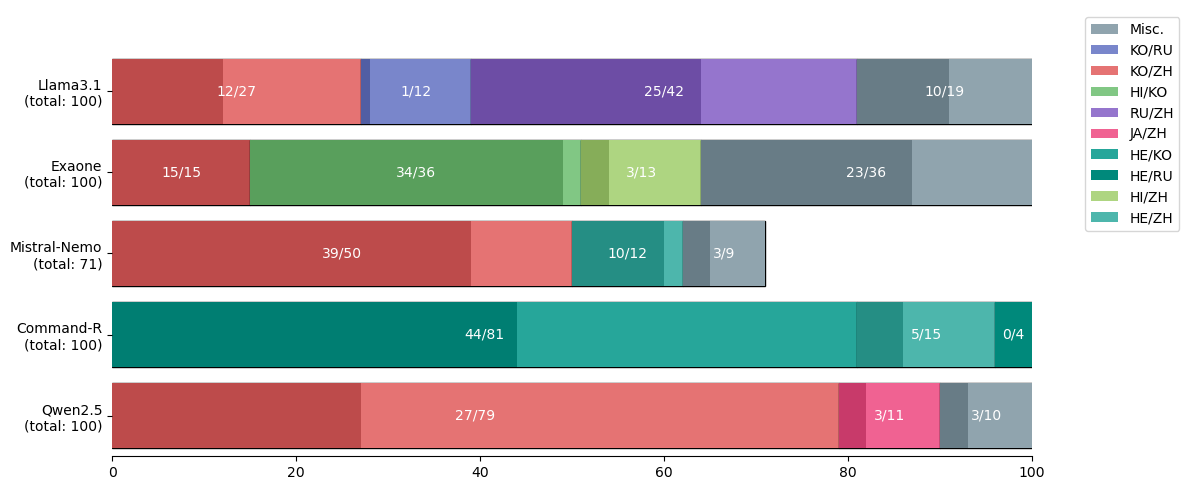}
        \caption{Improbable Bigrams}
        \label{fig:subfig_a}
    \end{subfigure}
    \vspace{1em}  
    \begin{subfigure}[b]{\textwidth}
        \centering
        \includegraphics[width=\textwidth]{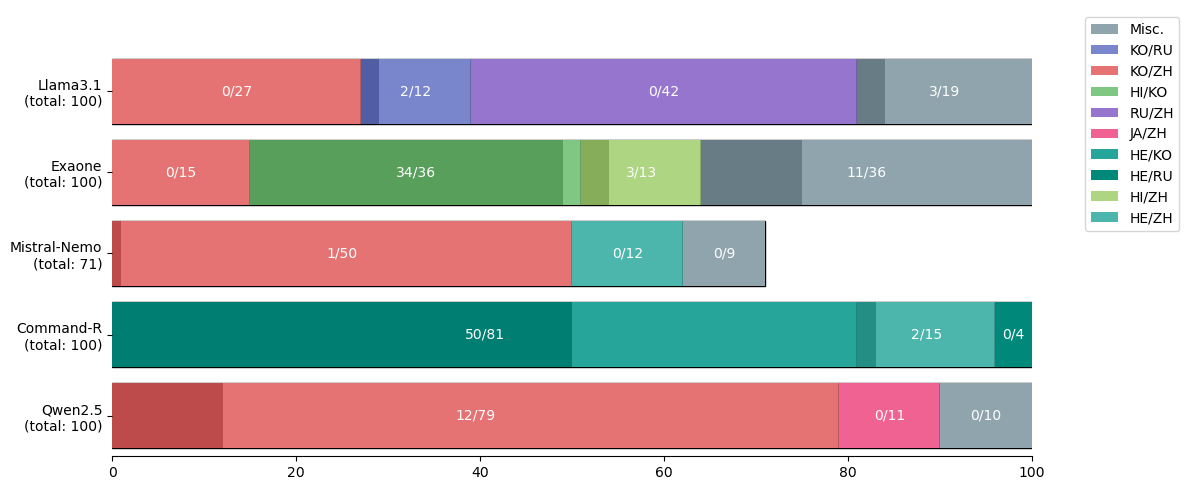}
        \caption{Improbable Bigrams (alternative tokenization)}
        \label{fig:subfig_b}
    \end{subfigure}
    \caption{Language-pair distributions of the improbable bigrams used in experiments. Darkened colors indicate improbable bigrams that cause hallucinations.}
    \label{fig:langs-bigrams}
\end{figure*}

\modified{
We analyze the scripts of the improbable bigrams used in the experiments.
Note that the language distributions of the tested improbable bigrams are not indicative of all possible improbable bigrams in each model's vocabulary, since the bigrams from the experiments were selected from the top half of the vocabulary when ordering by token trainedness (Section~\ref{sect: hall-exp}).
Figure~\ref{fig:langs-bigrams} shows the distribution of languages of the improbable bigrams for each model.
We observe that improbable bigrams can occur in a wide variety of languages, with higher resource multi-byte scripts such as Chinese, Korean, and Russian being the most frequent.
The diversity of languages of hallucinations suggest there is no significant influence caused by languages.
}

\modified{
Models had different distributions of language-pairs, which is influenced by both tokenizer training and model training.
For instance, Exaone had 17 unique language-pairs while Command-R had only 3 unique language-pairs.
Command-R's much larger tokenizer vocabulary may have impacted the incomplete token selection.
All of the 100 bigrams for Command-R contained Hebrew characters.
For comparison, there are no tokens in the top-half of Llama3.1's vocabulary that resolve to Hebrew characters.
Command-R is also the only model analyzed with bigrams of the Hebrew/Korean language-pair.
Bigrams of this language-pair constitute a majority of the tested improbable bigrams for Command-R, and performed worse when using alternative tokenization.
}

\noindent\textbf{Alternative Tokenization with Hebrew/Korean}
\
\modified{
The Command-R model's improbable bigrams did not show improvements when using alternative tokenization.
Figure~\ref{fig:langs-bigrams} shows that alternative tokenization harmed performance for the 81 Hebrew/Korean bigrams.
We investigate if the Hebrew/Korean language-pair is particularly prone to hallucinations when using alternative tokenization.
We collect a total of 38 possible bigrams of this language combination from the Llama3.1 vocabulary.
Repeating the experiments with Llama3.1's 38 Hebrew/Korean bigram pairs, using alternative tokenization reduced hallucinations (20/38) compared to regular tokenization (32/38).
These results suggest that Command-R's unusual result with alternative tokenization was not caused by the language-pairing of its improbable bigrams.
}

\end{document}